# AI Based Waste classifier with Thermo-Rapid Composting

Saswati kumari behera
*Department of Electrical and Electronics Engineering*
*Sri Sai Ram Engineering College*
Chennai, India
saswatikumari.eee@sairam.edu.in

Aouthithiye Barathwaj SR Y
*Department of Electrical and Electronics Engineering*
*Sri Sai Ram Engineering College*
Chennai, India
aouthithiyebarathwaj@ieee.org

Vasundhara L
*Department of Electrical and Electronics Engineering*
*Sri Sai Ram Engineering College*
Chennai, India
vasundhara@ieee.org

Saisudha G
*Department of Electrical and Electronics Engineering*
*Sri Sai Ram Engineering College*
Chennai, India
saisudhagurumoorthy@gmail.com

Haariharan N C
*Department of Electrical and Electronics Engineering*
*Sri Sai Ram Engineering College*
Chennai, India
haariharan@ieee.org

*Abstract*— Waste management is a certainly a very complex and difficult process especially in very large cities. It needs immense man power and also uses up other resources such as electricity and fuel. This creates a need to use a novel method with help of latest technologies. Here in this article we present a new waste classification technique using Computer Vision (CV) and deep learning (DL). To further improve waste classification ability, support machine vectors (SVM) are used. We also decompose the degradable waste with help of rapid composting. In this article we have mainly worked on segregation of municipal solid waste (MSW). For this model, we use YOLOv3 (You Only Look Once) a computer vision-based algorithm popularly used to detect objects which is developed based on Convolution Neural Networks (CNNs) which is a machine learning (ML) based tool. They are extensively used to extract features from a data especially image-oriented data.  In this article we propose a waste classification technique which will be faster and more efficient.  And we decompose the biodegradable waste by Berkley Method of composting (BKC).

## I. Introduction

Waste management process includes collection, storage, classification and transport of solid wastes. Waste management comprises all wastes such as industrial, household and biological wastes. All these process needs extensive planning, education and training. The wastes are mainly generated due to man-made activities. The solid waste is most common found such as vegetable waste, paper, plastics, wood, glass etc which are present especially in the urban areas in a large quantity. The classification of waste plays an important. The waste classification is tedious process especially in bigger cities. Waste management is very different all around the world i.e. each country has its own way of dealing with waste. It is very important for sustainable living. Proper management of waste is required for a clean, hygienic and an aesthetic environment. In current scenario people are employed manually to segregate waste (hand-picking method)[1]. This requires immense human labour. This method puts workers in danger as they segregate waste physically, they are exposed to the risk of infection by the microorganisms present in the solid waste. The conventional waste segregation is very expensive. The municipal solid waste (MSW) generated is mostly dumped in landfills and water bodies. The waste dumped as landfills pollutes both land and underground water. The underground water is a very important source for people especially in the crowded cities as it gets contaminated, it puts in risk the people who use the water for their daily uses. The landfill results in soil pollution mainly due to batteries and leaded substances. The municipal solid waste (MSW) generated alone is around ten thousand metric tons annually. These waste if not segregated and disposed carefully result as a hazard to the environment. The waste segregation should be done as early as possible in order to improve the percentage of recycled and reused materials and also to reduce the contamination caused by the litter. The waste segregation reduces the cost of disposal. Moreover, the waste segregation helps us to use appropriate waste disposal method and treatment method. About 75% of infectious disease is caused by improper disposal of waste, especially due to common household waste. This results in diseases like cholera, typhoid, tuberculosis etc as the waste gets mixed in the household used water and other components. On the other side when these mixed/ non segregated wastes burned can cause air pollution. Sometimes if these wastes contain any hazardous material when incinerated can cause respiratory oriented diseases in worst case it can cause cancer [2]. This mainly due to the fact that when these wastes are incinerated the minute hazardous particles are liberated out. This results in the formation of various respiratory oriented disordered. The field of computer vision and deep learning is growing day by day especially in field of waste management. Moreover, the image processing algorithms and architectures are growing rapidly which motivated us to coincide with the demand in the field of waste classification. The use of Convolutional neural networks in order to recognize and classify the captured image data. In order to classify the waste of different groups and type Support Vector Machines (SVMs) are used. This method is far more efficient and accurate when compared to the manual method of segregation i.e. hand-picking method [3].

## II. RESEARCH METHODOLOGY

### A. Classifers in practice

There are lot of methods which are in practice for the classification of the waste materials. The waste is classified with help of X-rays based on their density of the material. In trommel separator consists of a perforated rotating drum by which the waste is classified based on their size of the object. In this method larger wastes stays in the drum and smaller waste particle are collected separately. The common practice of waste classification in larger scale ais by Induction sorting. In this method the waste is placed and moved over a conveyor belt. Along the way of the conveyor belt there are number of sensors placed which detects the waste and classifies it accordingly like metal, wood, plastics etc. After that, these wastes are then moved to corresponding collector bin with the help of pressurized air jets which are placed along the conveyor belt. The next method of classification is based on electromagnetic theory. This segregator is known as Eddy current separator. This type of segregator is mainly used to separate metallic waste especially for separating ferrous from non-ferrous ones. This Eddy current separator is mainly used in metal-based industries for waste classification and unpopular for non-commercial usage. The other type and the most efficient one so far used in both commercial and non-commercial is Near-Infrared Sensors (NIR) [5]. In this method the waste objects are classified by their reflectance property. This method is the most effective among the so far discussed ways to classify waste [6]. As different materials possess different reflectance property the ability to distinguish and classification ability is large when compared to other discussed method. But if any waste overlays other this may give very uncertain results.

### B. Data Collection

In this field of image-based classification of objects the dataset plays a very important role. In order get the desired results to get to an appropriate conclusion on the supported fact data mining is very important. It is defined as the use of large dataset in an efficient manner with the help of mathematical and statistical tools. In image processing MATLAB, OpenCV are commonly used for this. The dataset consists of 470 images. The dataset is explicitly collected taking onto the account on the various factors involved while segregating waste. The factor includes mixed waste, overlying, irregular illuminance etc. The images are pre-processed in order to fit these irregularities. In this proposed model we use Support Vector Machine for data mining[7][8]. SVM can be used for both classification and regression problems. SVM plots its data in an n-dimensional space with its feature at a specific co-ordinate and its value. The classification is based on the separation of hyperplane depending upon its respective multidimensional data.

$$\min_{y,w,b} \frac{1}{2} \|w\|^2$$
$$y^{(i)}(w^T x^{(i)} + b) \geq 1, i = 1...m \quad (1)$$

The Eq (1) represents the optimisation of SVM where $y^i$ is the cost function, $w^T$ is the weights, $b$ is bias, $i$ is the example ranging from 1 to m. The given classification we use Linear SVM. As the features of the image dataset is higher that of the observations and for a higher dimensional data SVM are very effective. SVM reduces the memory consumption in image pre-processing. Also, in the decision function the use of various kernel functions is contributes to it strength.

## III. EXPERMINENT AND ANALYSIS

The proposed system consists of a Raspberry pi, a Raspberry pi high quality camera, a servo motor, a disc for auxiliary lid rotation, two compartments respectively for biodegradable and non-biodegradable wastes. In addition to this an compartment is added below this structure for rapid composting of biodegradable wastes.

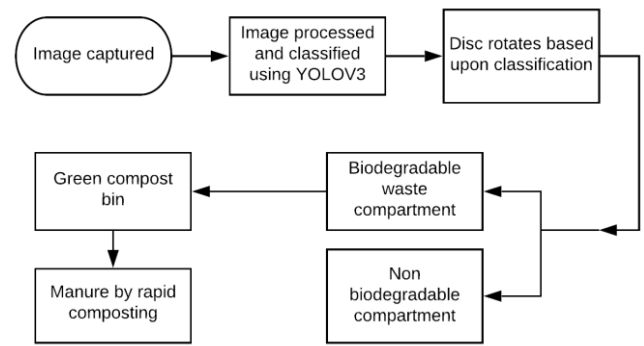

Fig. 1 The figure represents the flow of control and the data.

The waste when threw inside the bin it reaches the auxiliary lid under the preliminary opening. The waste is then captured by the Raspberry pi high quality camera. This camera has interchangeable lenses which helps us to collect high quality image data in low light conditions. As the image is captured inside the bin the illuminance inside is very low to overcome this issue the Raspberry pi high-quality camera is used with low-light lenses. The then captured is transmitted to the Raspberry pi where the image is processed and classified by using YOLOv3 weights which we trained using explicitly collected data [9-14]. The disc then rotates the auxiliary lid to the respective compartment depending upon their classes. The classified waste then reaches its respective compartment i.e. biodegradable and non-biodegradable waste compartment. The waste in the biodegradable compartment is then transferred to the rapid composting compartment. The composting is carried out based Berkley's method of rapid composting [15-19]. The conventional Berkley's method is altered in order to concise it in a waste bin. This novel composting depends upon thermal based composting. The compartment is 1 meter wide and 1.5 meter high, with a volume of 1.5 m³. The temperature of the compost is set around 55 - 65 °C. The C:N balance is to be maintained in the ratio of 30:1 in order to prevent decline in the rate of composting. The compost when set for the first is made wet with sufficient amount of water. Then an activator is added in between the wastes. The activator commonly used are animal urine, nettles, etc. The old compost can also be used as an activator and accelerator. The main function of activators is the inhibition of useful micro-organisms which helps in decomposing these waste. The moisture content and the C:N ratio helps the micro-organisms which act as decomposers to sustain and multiply This entire process takes up to 14 days for complete composting of the waste materials. This can not only dispose waste but also converts it into an useful product as it can be used as manure.

## IV. RESULTS

The pre-processed images are trained using YOLOv3 algorithm with SVMs. The training consists of 470 images of 2 classes the biodegradable and the non-biodegradable. The labeling of the dataset is done by using LABELIMG. The training and the test dataset are split in 9:1 ratio. The training is done in Google Colaboratory. The MAP (Mean Average Precision) is calculated for every 100 iterations and the maximum batch is set at 2500.

The results so obtained is much efficient and accurate when compared to training the data without image pre-processing when trained on YOLOv3 algorithm. The image preprocessing with the help of Support machine vectors helps us not only increase the accuracy of detection but also it reduces the memory utilization.

**Table 1** The table shows the mineral values from the compost

| S.No. | Decomposed waste contents | |
|---|---|---|
| | *Component* | *Percentage* |
| 1 | Organic matter | 80.92 |
| 2 | Organic Carbon Content | 30.75 |
| 3 | Nitrogen Content | 1.3856 |
| 4 | Ash Content | 30.75 |

The table 1, gives the percentage of respective organic matter, carbon content, nitrogen and ash content after 14 days of decomposition. This novel method yields better carbon and nitrogen content when compared with cold composting. Moreover, in this method disease causing microorganisms are eliminated.

## V. CONCLUSION

In the proposed system we were able to increase the accuracy and the efficiency of the waste classification with the help of computer vision and deep learning. This reduces the dependency in the manual labor and reduces the cost also. The novel composting of the biodegradable waste based on Berkley Method of rapid composting, yields a better result when compared to the conventional method. By reducing the size needed for classification and decomposition of biodegradable wastes.

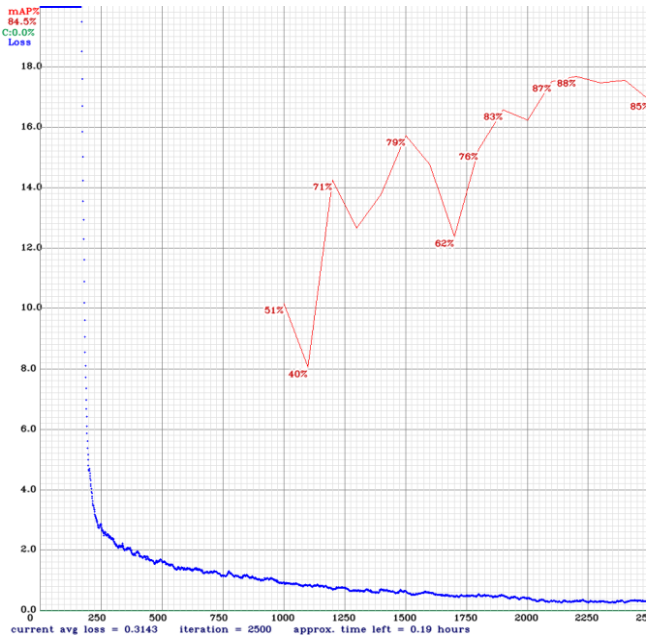
Fig. 2 The diagram shows change of MAP for 2500 iterations

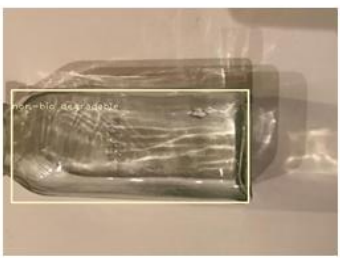
(a)

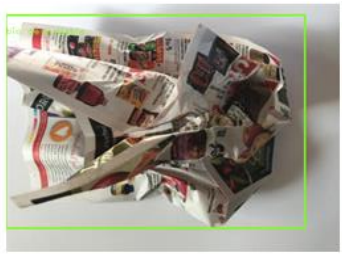
(b)

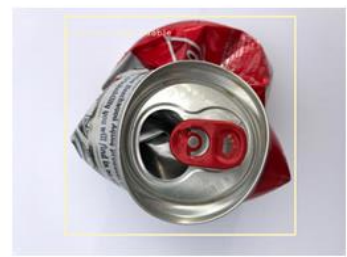
(c)

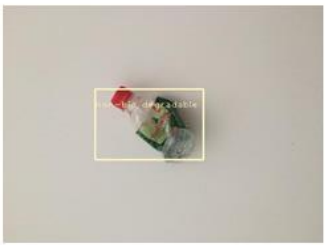
(d)

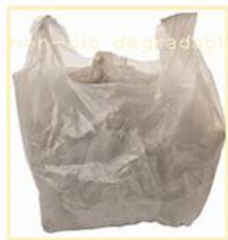
(e)

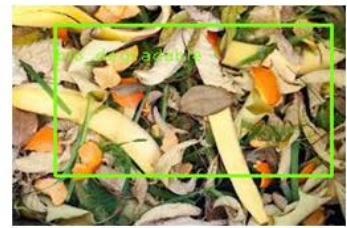
(f)

Fig. 3 The images (a), (c), (d), (e) shows the detection of non-biodegradable waste and (b), (f) shows the biodegradable waste.